\documentclass[journal,10pt,twoside]{IEEEtran}
\usepackage{amsmath,amsfonts}
\usepackage{algorithmic}
\usepackage{array}
\usepackage[caption=false,font=normalsize,labelfont=rm,textfont=rm]{subfig}
\usepackage{textcomp}
\usepackage{stfloats}
\usepackage{url}
\usepackage{verbatim}
\usepackage{graphicx}
\def\BibTeX{{\rm B\kern-.05em{\sc i\kern-.025em b}\kern-.08em
		T\kern-.1667em\lower.7ex\hbox{E}\kern-.125emX}}
\usepackage{balance}
\usepackage[utf8]{inputenc}
\usepackage[ruled,vlined]{algorithm2e}
\usepackage{hyperref}
\usepackage{makecell}
\hypersetup{
	colorlinks = true, 
	linkcolor = blue,
	urlcolor = blue,
	citecolor = blue
}
\usepackage{color}

\usepackage{graphicx}
\usepackage{epstopdf}
\begin{document}
	\title{Deep-Reinforcement-Learning-Based AoI-Aware Resource Allocation for RIS-Aided IoV Networks}

	\author{Kangwei Qi, Qiong Wu,~\IEEEmembership{Senior Member,~IEEE}, Pingyi Fan,~\IEEEmembership{Senior Member,~IEEE}, \\Nan Cheng,~\IEEEmembership{Senior Member,~IEEE}, Wen Chen,~\IEEEmembership{Senior Member,~IEEE}, \\Jiangzhou Wang,~\IEEEmembership{Fellow,~IEEE} and Khaled B. Letaief,~\IEEEmembership{Fellow,~IEEE}
		\thanks{This work was supported in part by the National Natural Science Foundation of China under Grant No. 61701197, in part by the National Key Research and Development Program of China under Grant No.2021YFA1000500(4), in part by the National key project 2020YFB1807700, NSFC 62071296, Shanghai Kewei 22JC1404000, in part by the Research Grants Council under the Areas of Excellence scheme grant AoE/E-601/22-R, in part by the 111 Project under Grant No. B12018. (Corresponding authors: Qiong Wu.)
			
		Kangwei Qi, Qiong Wu are with the School of Internet of Things Engineering, Jiangnan University, Wuxi 214122, China. (e-mail: kangweiqi@stu.jiangnan.edu.cn, qiongwu@jiangnan.edu.cn).
			
		Pingyi Fan is with the Department of Electronic Engineering, Beijing National Research Center for Information Science and Technology, Tsinghua University, Beijing 100084, China (e-mail: fpy@tsinghua.edu.cn).
			
		Nan Cheng is with the State Key Lab. of ISN and School of Telecommunications Engineering, Xidian University, Xi’an 710071, China (e-mail: dr.nan.cheng@ieee.org).
			
		Wen Chen is with the Department of Electronic Engineering, Shanghai Jiao Tong University, Shanghai 200240, China (e-mail: wenchen@sjtu.edu.cn).
			
		Jiangzhou Wang is with the School of Engineering, University of Kent, CT2 7NT Canterbury, U.K. (email: j.z.wang@kent.ac.uk).
			
		K. B. Letaief is with the Department of Electrical and Computer Engineering, the Hong Kong University of Science and Technology (HKUST), Hong Kong (email:eekhaled@ust.hk).
			
		}
	}
	
	\maketitle
	
	\begin{abstract}
		Reconfigurable intelligent surface (RIS) is a pivotal technology in communication, offering an alternative path that significantly enhances the link quality in wireless communication environments. In this paper, we propose a RIS-assisted internet of vehicles (IoV) network and consider the vehicle-to-everything (V2X) communication method. In order to improve the timeliness of vehicle-to-infrastructure (V2I) links and the stability of vehicle-to-vehicle (V2V) links, we introduce the age of information (AoI) and the payload transmission probability models. Thus, we aim to minimize the AoI of V2I links and prioritize transmission of V2V links payload. In this framework, the base station (BS) server acts as the agent responsible for resource allocation for vehicles and controlling the phase-shift of the RIS. We use the Soft Actor-Critic (SAC) algorithm to address this problem due to its gradual convergence and high stability in the optimization process. An AoI-aware joint vehicular resource allocation and RIS phase-shift control scheme based on SAC algorithm is proposed and simulation results show that its convergence speed, cumulative reward, AoI performance, and payload transmission probability outperform those of proximal policy optimization (PPO), deep deterministic policy gradient (DDPG), twin delayed deep deterministic policy gradient (TD3) and stochastic algorithms.
	\end{abstract}
	

	\section{Introduction}
	\IEEEPARstart{I}{n} recent years, with the rapid advancement of internet of things (IoT), vehicles have undergone a significant transformation towards increased intelligence \cite{wu1008, wu1012}. This has led to a higher demand for vehicle communication technology, prompting numerous organizations to engage in in-depth research to meet diverse requirements in internet of vehicles (IoV) networks \cite{r1, zhang1, wu1002, wu1010}. Vehicle-to-everything communication (V2X), as an important technology, has brought multiple enhancements to vehicles in communication \cite{wu1013, wu1014}. V2X technology covers a variety of scenarios, including vehicle-to-vehicle communication (V2V), vehicle-to-infrastructure communication (V2I), vehicle-to-pedestrian communication (V2P), and vehicle-to-network communication (V2N) \cite{r2, cheng1}. This diversity of communication scenarios provide rich possibilities for different applications. For example, V2I communication can enable vehicles to obtain road conditions, traffic signal conditions, driving routes from base station (BS), V2V communication enables vehicles to exchange key information such as position, speed, and acceleration with each other \cite{r3, r5, wu1011}.
	
	Enhancements in V2X technology have made vehicle communications faster, more reliable, and more diverse, providing important support for smart driving and intelligent transportation \cite{r6, wu1007}. In order to better increase the transmission rate of the V2I links and ensure reliable delivery of information in V2V links, we consider the resource allocation in the vehicle network. An effective resource allocation scheme can increase the information transmission rate V2I links and ensure reliable transmission of V2V links \cite{r8, wu1003, wu1009}. In addition, the high mobility of vehicles also has an impact on vehicular communication, and thus it becomes challenging to realize an efficient resource allocation scheme \cite{wu1}. Fortunately, the continuing development of artificial intelligence has resulted in deep reinforcement learning (DRL), which combines deep learning (DL) and reinforcement learning (RL) \cite{wu2, cheng2, wu1015}. DRL is proving to be a more promising approach in autonomous decision making and vehicle control \cite{xiong3, wu1004}. DRL addresses resource allocation problems by formulating them as Markov decision process (MDP), where deep neural networks (DNNs) are able to automatically learn higher-level feature representations from raw data through multi-level non-linear transformations \cite{wu4, wu1005, wu1006}.
	
	Nowadays, more and more researches adopt DRL to solve various problems in V2X communication \cite{shao1}. In \cite{r12}, Liang \emph{et al.} delved into the spectrum sharing problem within vehicular environments, by employing a multi-agent deep reinforcement learning (MADRL) approach. Their focus was on enhancing the sum capacity of V2I links and the payload transmission success rate of V2V links. In \cite{r13}, Yuan \emph{et al.} employed the deep Q-network (DQN) algorithm for resource blocks (RBs) allocation and deep deterministic policy gradient (DDPG) for transmission power control. To enhance adaptability in dynamic environments, the authors introduced a meta-based DRL algorithm, amalgamating meta-learning and DRL.
	In \cite{r14}, Ye \emph{et al.} proposed a distributed RL algorithm for resource allocation in unicast and broadcast scenarios involving hybrid V2V and V2I communication. Their aim was to minimize V2I communication interference while meeting V2V constraints, allowing each agent to make decisions without requiring global information. Additionally, in \cite{r15} and \cite{r16}, Yang et al. and Zhao et al. considered three communication modes: reuse, dedicated, and cellular. Yang et al. introduced the efficient transfer actor-critic (ETAC) learning algorithm to optimize mode selection, RBs allocation, and power control for vehicles. Their objective was to maximize the sum capacity of V2I and V2V links while ensuring the ultra-reliable low-latency communication (URLLC) requirements for V2V links. Meanwhile, Zhao et al. established an optimal link scheme between vehicles to ensure V2V communication reliability. They utilized each V2V link as an agent to optimize mode selection and power control through the double DQN (DDQN) algorithm. These above researches have solved some problems such as resource allocation in vehicular network by DRL method, which increases the information transmission rate of the V2I link and ensures the transmission of secure information in the V2V link. 
	
	However, in the design of intelligent transport systems (ITS), both the age of information (AoI) of the V2I link and the reliable transmission of the V2V link must be considered \cite{r17}. In practical applications, such as in the communication between self-driving cars and traffic signals, a lower AoI ensures that the vehicle can obtain the state change of the signal in a timely manner, which optimizes the driving route, reduces the waiting time and improves the traffic flow. Meanwhile, in an emergency, reliable transmission over the V2V link enables one vehicle to quickly send brake warnings to surrounding vehicles, reducing the risk of collision. In addition, during convoy driving, real-time AoI updates can help vehicles coordinate their speed and position for more efficient collective movement \cite{r18}. These applications show that integrating both is critical to improving the safety and efficiency of ITS.
	In \cite{r19}, Parvini \emph{et al.} designed a multi-agent DDPG (MADDPG) resource allocation algorithm with task decomposition to solve a series of optimization problems such as the selection of transmission modes, RB allocation, and power control for vehicular platforms, using multiple vehicles as a platform to minimize the AoI of the platforms while ensuring that security information is transmitted to the platform members. In \cite{r20}, Peng \emph{et al.} established an AoI model about V2I and V2V, and proposed an AoI-aware joint spectrum and power dynamic allocation scheme based on the trust region policy optimization (TRPO) algorithm with the objective of minimizing the average AoI of all V2I links and the average power consumption of all V2V links. In \cite{r21}, Mlika \emph{et al.} investigated the AoI minimization problem in the V2X ecosystems, which includes coverage optimization, half-duplex transceiver selection, power allocation, and RBs scheduling by applying non-orthogonal multiple access (NOMA) techniques, in order to solve this hybrid problem containing discrete and continuous actions, the authors propose a decomposition-based greedy matching and DDPG algorithm, which solves the RBs scheduling and transceiver selection problems by a roommate matching algorithm, and then solves the power allocation and coverage optimization problems by DDPG.
	In \cite{r8}, Liang \emph{et al.} designed a resource allocation scheme based on slowly varying large-scale fading information to maximize total V2I transmission links capacity and the minimum capacity in V2I links while ensuring V2V transmission reliability. In \cite{r9}, Guo \emph{et al.} considered packet retransmission and analyzed the M/G/1 queuing model of V2V to ensure the probability of packet loss as well as the average packet dwell time for V2V. In addition, in \cite{r10}, Guo \emph{et al.} obtained the steady-state reliability and delay expressions for queue-based spectrum reuse by analyzing the V2V queue model. In \cite{r11}, Mei \emph{et al.} modeled the V2V packet arrival process as a Poisson distribution to further analyze the delay and reliability of V2V communication.
	
	When the vehicle is traveling, it usually encounters obstacles to impede communication, which will reduce the vehicle's information transmission rate to a certain extent, affecting the vehicle's AoI and making the vehicle unable to receive the most recently updated road traffic conditions in a timely manner. Therefore, we will use reconfigurable intelligent surface (RIS) to assist vehicle communication. Vehicles can send signals to the RIS, which reflects and spreads the signals to other vehicles or infrastructure \cite{xiong1, xiong2}. In addition, RIS is considered as an intelligent thin composite material similar to wallpaper that covers parts of walls, buildings, ceilings, etc., with a high degree of flexibility \cite{r22, r23, wu1001}. The RIS is a passive array structure capable of adjusting the phase-shift of each reflective element on the surface either almost continuously or discretely, so by training to obtain the optimal phase-shift, it can increase the vehicle's information transmission rate and reduce the vehicle's AoI \cite{r24, r25}. 
	
	Due to the above and other advantages of RIS, this has resulted in a large number of related researches, in \cite{r26} Guo \emph{et al.} explored the use of RIS to optimize AoI in mmWave communication systems. By jointly optimizing beamforming, RIS reflection coefficients and user scheduling strategies, the algorithm proposed in the article meets the freshness of information requirement while increasing the total system rate. In \cite{r27}, Muhammad \emph{et al.} proposed a scheme based on RIS and Cooperative Non-Orthogonal Multiple Access (C-NOMA), which effectively reduces the AoI in real-time IoT applications by optimizing the transmission power and RIS phase-shift matrix. In \cite{r28}, Sun \emph{et al.} presented a three-step scheme to simultaneously reduce the AoI and increase the transmission data rate in unmanned aerial vehicle (UAV) assisted IoT networks by jointly optimizing the UAV position, RIS phase-shift matrix and transmission scheduling.
	In \cite{r29}, Huang \emph{et al.} designed a joint BS transmission beam matrix and RIS phase-shift optimization to maximize the sum of multi-user downlink rates via the DDPG algorithm. In \cite{r30}, Ji \emph{et al.} considered a device-to-device (D2D) communication where the RB allocation and power control  for each user is first trained by a multi-agent reinforcement learning algorithm, and then a centralized DDQN algorithm is used to optimize the RIS phase-shift and position to maximize the sum of information rates. In \cite{r31}, Gu \emph{et al.} investigated the joint power control and phase-shift design in RIS-assisted vehicular communication to maximize the V2I capacity while guaranteeing low-latency and high-reliability of the V2V link, and in \cite{r32}, Chen \emph{et al.} considered multi vehicular users, and in order to guarantee the quality-of-service (QoS) requirements of V2I and V2V, the transmission power of the vehicle, the multi-user detection (MUD) matrix, spectrum reuse of V2V links and RIS reflection coefficients were jointly optimized.
	
	Reviewing the above related studies, we find that although there have been many advances in RIS-assisted vehicular networks, comprehensive studies that simultaneously consider channel sharing, power allocation, and phase-shift control are still relatively lacking. This is mainly due to a number of challenges in RIS-assisted vehicular communication. Firstly, the scenarios are complex and involve the dynamics of multiple vehicles. Second, vehicles move at high speeds resulting in real-time changes in their positions, which affects the channel gain between links. Third, the timeliness of V2I links and the reliability of V2V links also need to be considered simultaneously. Finally, the three problems of channel sharing, power allocation and phase-shift control are considered together to form a non-convex optimization problem, which increases the complexity of the study.
	
	In this paper, we investigate the resource optimization and phase-shift matrix design problem for RIS-assisted vehicular communications in a high-mobility vehicular environment, so as to minimize the sum AoI of all V2I links as well as to increase the payload transmission success rate of V2V links. The main contributions of this work are summarized as follows\footnote{The source code has been released at: https://github.com/qiongwu86/RIS-RB-AoI-V2X-DRL.git}:
	
	\begin{itemize}
		\item [1)]Our focus is on optimizing RIS-assisted vehicular communication to enhance AoI in V2I links and the reliability of V2V transmission payloads. To address this, we formulate the long-term optimization problem as a MDP.
		
		\item [2)]Given the independent nature of RIS, the lack of shared phase-shift information among vehicle users poses a challenge. We propose a solution where the BS takes on the responsibility of allocating channel resources, determining transmission power for all vehicle users, and deciding the RIS phase-shift matrix. Leveraging the DRL framework, we devise an AoI-aware joint vehicular resource allocation and RIS phase-shift control scheme based on the SAC algorithm. This approach aims to improve the performance of both V2I and V2V links.
		
		\item [3)]We validate the effectiveness of our algorithm through simulation experiments, comparing it with DDPG, proximal policy optimization (PPO), twin delayed deep deterministic policy gradient (TD3), and various stochastic algorithms. By varying parameters, we demonstrate the algorithm's advantages in terms of convergence, reward performance, and stability.
	\end{itemize}
	
	The rest of this paper is organized as follows. Section II describes the system model briefly. Section III sets up the MDP framework to model resources allocation in vehicular networks and solves the problem by SAC algorithm. Section IV presents the simulation results. Section V concludes this paper. 
	
	\section{System Model}
	\begin{figure*}[t]
		\centering
		\includegraphics[width=5in]{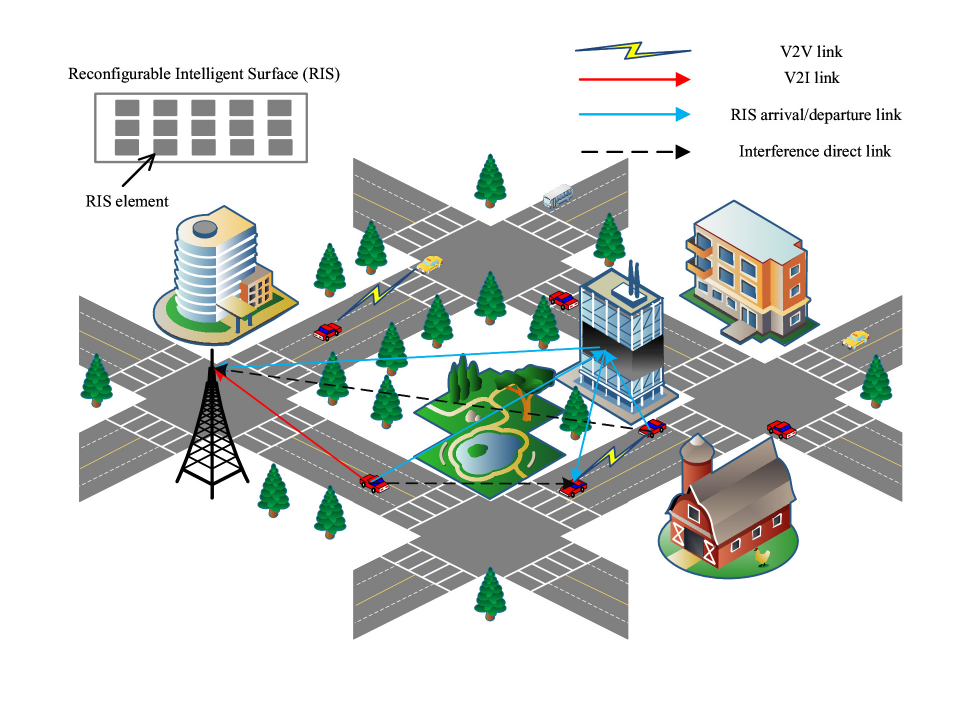}
		\vspace{-0.5cm}
		\caption{Reconfigurable intelligent surface aided IoV networks}
		\label{fig1}
	\end{figure*}
	
	\subsection{Scenario}
	As shown in Fig. \ref{fig1}, consider a vehicular communication network consisting of RIS-assisted V2I and V2V links, where the vehicles are randomly distributed on a 450$*$650 square meter road network, there are paths up and down, left and right, with three double lanes of road in each direction. When encountering an intersection, there is a certain probability that the vehicle will choose to turn or go straight ahead, and furthermore, we consider that there are $M$ vehicles, each of which plays the role of both a cellular vehicular user (CUE) and a device-to-device vehicular user (DUE) in the system, and thus there are $\mathit{M}$ CUEs, which communicate with the BS by V2I link (we consider an uplink scenario, where the vehicle is equipped with an antenna and the BS is equipped with an $\mathit{M}$-element uniform linear array), and $\mathit{K}$ pairs of DUEs , which communicate directly with each other via V2V link. The sets of CUEs and DUEs are defined as $\mathcal{M}=\left\{1,2,\cdots,\mathit{M}\right\}$ and $\mathcal{K}=\left\{1,2,\cdots,\mathit{K}\right\}$, respectively. Furthermore, assuming that the $\mathit{M}$ V2I links have been pre-allocated orthogonal spectrum with fixed transmission power, each DUE needs to reuse the spectrum of a CUE, denoted by $\mathit{x_{m,k}^n}$ spectrum sharing case, and $\mathit{x_{m,k}^n}=1$ indicates that the $\mathit{k}$th DUE reuses the spectrum of the $\mathit{m}$th CUE at slot $n$. Different types of V2X links interfere with each other, and in order to improve the quality of service of the communication link, RIS is deployed to enhance the performance of the communication system, where RIS is a uniform rectangular array of $\mathit{F}$ passive elements, taking the V2I link as an example, the vehicle can communicate with the BS directly or with the BS through RIS assistance, and we assume that both communication links exist simultaneously.
	
	\subsection{Mobility Model}
	At slot $\mathit{n}$, we assume that the coordinates of the $\mathit{m}$th vehicle is $(x_m^n,y_m^n,z_m^n)$, the speed of is $v_m^n$. Here we take the example of a vehicle traveling upward (the y-axis direction is up), at an intersection, when a vehicle chooses to continue straight, its position is constant in the direction of the x-axis and changes in the direction of the y-axis, then the position change can be represented as 
	\begin{equation}\label{eq1}
		y_m^{n+1} = y_m^n + v_m * {\tau_0},
	\end{equation} 
	where $\tau_0$ is slow fading time slot interval, when a vehicle chooses to turn left, its position change can be represented as 
	\begin{equation}\label{eq2}
		\left\{ {\begin{array}{*{20}{c}}
				{x_m^{n+1} = x_m^n - (v_m * {\tau_0} - \Delta {\rm{distance)}}},\\
				{y_m^{n+1} = \emph{{coordinate of the left-turning road}}},
		\end{array}} \right.
	\end{equation}
	where $\Delta \rm{distance} $ is the distance traveled by the vehicle just before it turned. When a vehicle chooses to turn right, its position change can be represented as 
	\begin{equation}\label{eq3}
		\left\{ {\begin{array}{*{20}{c}}
				{x_m^{n+1} = x_m^n + (v_m * {\tau_0} - \Delta {\rm{distance)}}},\\
				{y_m^{n+1} = \emph{{coordinate of the right-turning road}}},
		\end{array}} \right.
	\end{equation}

	\subsection{Communication Model}
	It is assume that the coordinates of the BS and the RIS are
	$({x_{BS}},{y_{BS}},{z_{BS}})$ and $({x_{RIS}},{y_{RIS}},{z_{RIS}})$, respectively. At slot $\mathit{n}$, the coordinates of the transmitter in the $\mathit{k}$th D2D pair is
	$(x_{k,n}^{{D_t}},y_{k,n}^{{D_t}},z_{k,n}^{{D_t}})$, the coordinates of the receiver is 
	$(x_{k,n}^{{D_r}},y_{k,n}^{{D_r}},z_{k,n}^{{D_r}})$.
	In addition, the set of available channels is defined as $\mathcal{L}=\left\{1,2,\cdots,\mathit{L}\right\}$, and according to \cite{r9}, the channel gain of the $\mathit{k}$th V2V link occupying the $\mathit{l}$th sub-channel at slot $\mathit{n}$ is 
	
	\begin{equation}\label{eq4}
		h_k^n[l] = \sqrt {\rho \beta {{(d_k^n)}^{ \eta  }}} g_k^n[l],
	\end{equation}
	where $\rho $ is the path loss constant for the reference distance $\mathit{d_{0}} = 1\mathit{m}$, $\beta$ is the logarithmic shaded fading random variable, $d_k^n$ is the geometric distance between the transmitter and the receiver in the $\mathit{k}$th V2V pair at slot $\mathit{n}$, and $\eta$ is the corresponding path loss exponent. $g_k^n[l]$ is the frequency-dependent small-scale fading power component, which is assumed to be exponentially distributed with unit mean. Assuming that all are communicating on the $\mathit{l}$th channel at slot $\mathit{n}$, the channel coefficients from the $\mathit{m}$th CUE to the BS, $h_{m,b}^n[l]$, the channel coefficients from the $m$th CUE to the receiver in the $k$th V2V pair, $h_{m,k}^n[l]$, the channel coefficients from the $\mathit{k}'$th V2V transmitter to the $\mathit{k}$th V2V receiver, $h_{k',k}^n[l]$, and the channel coefficients from the $k$th V2V transmitter to the BS, 
	$h_{k,b}^n[l]$, are all defined in a similar manner as in the above equation.
	
	The channel gain from the $\mathit{m}$th CUE to the RIS is defined as 
	\begin{equation}\label{eq5}
		h_{m,r}^n[l] = \sqrt {\rho \beta {{(d_{m,r}^n)}^{ - \eta }}} {e^{ - j2\pi \frac{{d_{m,r}^n}}{{\lambda [l]}}}}h_{AoA}^n[l],
	\end{equation}
	where the arrival array response of the RIS can be defined as 
	\begin{equation}\label{eq6}
		h_{A{\rm{o}}A}^n[l] = {\left[ {1, \cdots ,{e^{ - j2\pi \frac{d}{{\lambda [l]}}(N - 1)\sin (\theta _{AoA}^n)}}} \right]^{\rm T}},
	\end{equation}
	where $\mathit{d}$ is the spacing between each element in the RIS, and 
	$\theta _{AoA}^n$ is the angle of arrival (AoA) of a signal at slot $\mathit{n}$. By the same way, we can express the channel gain from the RIS to the BS at slot $\mathit{n}$ as 
	$h_{r,b}^n[l]$, the channel gain from the transmitter in the $\mathit{k}$th V2V pair to the RIS as $h_{k,r}^n[l]$ 
	and the channel gain from the RIS to the receiver in the $\mathit{k}$th V2V pair as $h_{r,k}^n[l]$. 
	
	Therefore, at slot $\mathit{n}$, the received Signal-to-Interference-plus-Noise Ratio (SINR) for the $\mathit{m}$th V2I link and the $\mathit{k}$th V2V link on the $\mathit{l}$th sub-channel is expressed as
	\begin{equation}\label{eq7}
		\gamma _m^n[l] = \frac{{P_m^c|{{(h_{r,b}^n[l])}^{\rm H}}{\Theta ^n}h_{m,r}^n[l] + h_{m,b}^n[l]{|^2}}}{{\sum\limits_{k \in K} {x_{m,k}^nP_{k,n}^d|({{(h_{r,b}^n[l])}^{\rm H}}{\Theta ^n}h_{k,r}^n[l] + h_{k,b}^n[l]){|^2} + {\sigma ^2}} }},
	\end{equation}
	\begin{equation}\label{eq8}
		\gamma _k^n[l] = \frac{{P_{k,n}^d|{{(h_{r,k}^n[l])}^{\rm H}}{\Theta ^n}h_{k,r}^n[l] + h_k^n[l]{|^2}}}{{I_k^n[l] + {\sigma ^2}}},
	\end{equation}
	where $\mathit{P_m^c}$ denotes the V2I transmit  power, $P_{k,n}^d$ denotes the V2V transmit  power at slot $\mathit{n}$, the diagonal phase-shift matrix of RIS 
	${\Theta ^n} = diag[{\beta _1}{e^{j\theta _1^n}}, \cdots ,{\beta _f}{e^{j\theta _f^n}}, \cdots ,{\beta _F}{e^{j\theta _F^n}}],\forall f \in [1,F]$, due to the hardware limitations, the phase-shift can only be selected from a finite set of discrete values, where 
	$\theta _f^n \in \left\{ {0,\frac{{2\pi }}{Q}, \cdots ,\frac{{2\pi (Q - 1)}}{Q}} \right\}$ and 
	${\beta _f} \in [0,1]$ are used as associated transmission coefficients and amplitudes of the phase-shift, where $Q$ is the quantity that controls the degree of phase-shift discrete, and
	\begin{equation}\label{eq9}
		\begin{array}{l}
			\begin{aligned}
				I_k^n[l] =& P_m^c{\left| {{{(h_{r,k}^n)}^{\rm H}}{\Theta ^n}h_{m,r}^n[l] + h_{m,k}^n[l]} \right|^2} + \\
				&\sum\limits_{m \in M} {x_{m,k'}^nP_{k',n}^d{{\left| {h_{k',r}^n[l]{\Theta ^n}h_{r,k}^n[l] + h_{k',k}^n[l]} \right|}^2}}.
			\end{aligned} 
		\end{array}
	\end{equation}
	
	Then, the information transmission rate of the $\mathit{m}$th CUE and the information transmission rate of the $\mathit{k}$th V2V pair are expressed as:
	\begin{equation}\label{eq10}
		R_m^n[l] = W{\log _2}(1 + \gamma _m^n[l]),
	\end{equation}
	and
	\begin{equation}\label{eq11}
		R_k^n[l] = W{\log _2}(1 + \gamma _k^n[l]).
	\end{equation}
	where $\mathit{W}$ is the bandwidth of the spectrum sub-band.
	
	\subsection{AoI and Probability Model}
	As previously described, the V2I communication must exchange information with the BS in time, and 
	$R_m^n[l]$ must be strictly greater than or equal to $R^{th}$, which causes the AoI to reset to 
	$\Delta n$ to ensure the freshness of the information, where $\Delta n$ denotes the channel coherence time, $R^{th}$ is denoted as the minimum rate required for the V2I communication and the AoI of V2I is increasing when $R_m^n[l]$ is less than $R^{th}$. Here, we define the AoI of the $m$th V2I communication as 
	$A_m^n$, which denotes the time elapsed since the most recent successful communication \cite{r34}. The AoI evolves through the following equation
	\begin{equation}\label{eq12}
		A_m^{n + 1} = \left\{ {\begin{array}{*{20}{c}}
				\Delta n,\\
				{A_m^n + \Delta n,}
		\end{array}} \right.\begin{array}{*{20}{c}}
			{{\rm{if\quad}}R_m^n \ge {R^{th}},}\\
			{{\rm{otherwise,}}}
		\end{array}
	\end{equation}
	
	Meanwhile, to ensure the reliability of V2V transmission of security information, we model the probability that V2V transmission will successfully deliver a packet of size $D$ within a time budget $N$ as \cite{r12}
	\begin{equation}\label{eq13}
		\Pr \left\{ {\sum\limits_{n = 1}^N {\sum\limits_{l = 1}^L {x_{m,k}^n[l]} }  \cdot R_k^n[l] \cdot \Delta n \ge D} \right\},
	\end{equation}
	where $D$ denotes the size of the periodically generated V2V payload.
	
	\subsection{Problem Formulation}
	In order to improve the relevant performance of the communication system, we investigate a framework for optimizing the channel resource allocation, power control, and RIS phase-shift matrix, which is formulated as an optimization problem whose objective is to minimize the AoI of the V2I links and maximize the transmission success of the V2V payloads, whereupon, the problem can be expressed as
	\begin{subequations}\label{P1}
		\begin{equation}\label{eq14a}
			P1:{\rm{}}\mathop {\min }\limits_{\{ \mathbf{X},\mathbf{P},{\mathbf{\Theta}}\}} \left\{ \begin{array}{l}
				\frac{1}{N}\sum\limits_{n = 1}^N {A_m^n}  - \\
				\Pr \left\{ {\sum\limits_{n = 1}^N {\sum\limits_{l = 1}^L {{x_{m,k}^n}[l]} }  W R_k^n[l]\Delta n \ge D} \right\}
			\end{array} \right\},
		\end{equation}
		\begin{equation}\label{eq14b}
			{\rm{               }}s.t.\quad{\rm{  }}R_{m,k}^n[l] \ge {R^{th}},{\rm{    }}\forall m \in M,\forall k \in k,
		\end{equation}
		\begin{equation}\label{eq14c}
			\qquad\qquad\qquad{\rm{                     }}0 < P_{k,n}^d < P_{\max }^d,{\rm{  }}\forall k \in K,{\rm{ }}\forall n \in N,
		\end{equation}
		\begin{equation}\label{eq14d}
			\quad\qquad\qquad{\rm{                     }}x_{m,k}^n \in \left\{ {0,1} \right\},{\rm{       }}\forall m \in M,{\rm{ }}\forall k \in K,
		\end{equation}
		\begin{equation}\label{eq14e}
			{\rm{                    }}\sum\limits_{\rm{m}} {x_{m,k}^n \le 1{\rm{,         }}\forall m \in M,} {\rm{ }}\qquad
		\end{equation}
		\begin{equation}\label{eq14f}
			\qquad{\rm{                    0}} \le \theta _f^n \le 2\pi ,{\rm{       }}\forall f \in F,\forall n \in N,
		\end{equation}
	\end{subequations}
	the constraint (\ref{eq14c}) imposes limitations on the transmission power range within a V2V pair. Constraints (\ref{eq14d}) and (\ref{eq14e}) assume that each V2V pair occupies only one channel. Additionally, constraint (\ref{eq14f}) restricts the range of RIS phase-shift, designed to be discrete due to hardware limitations. The objective function is non-convex, which is a challenge for traditional optimization algorithms as they struggle to find a solution very quickly. Consequently, in the next section, we will explore the application of DRL methods to address and solve this non-convex problem.
	
	\section{DRL-Based Problem Solving}
	The optimization objective $P$1 involves vehicle resource allocation and RIS phase-shift control based on AoI sensing. To address this problem, we formulate it as a DRL process. In this approach, the central BS handles the centralization of vehicle channel selection, power allocation, and RIS phase-shift matrix control. At each slot $n$, the agent observes the current state $s_{n}$, subsequently taking an action $a_n$
	based on the current policy. Following this action, the agent receives relevant rewards $r_n$
	from the system environment. The state then transitions to the next state, marking the completion of a cycle in this dynamic process. This entire sequence can be succinctly represented as ${e_n} = ({s_n},{a_n},{r_n},{s_{n + 1}})$, encapsulating the essential components of the DRL framework for solving the optimization problem.
	
	\begin{figure*}[t]
		\centering
		\vspace{-0.4cm}
		\includegraphics[width=6in]{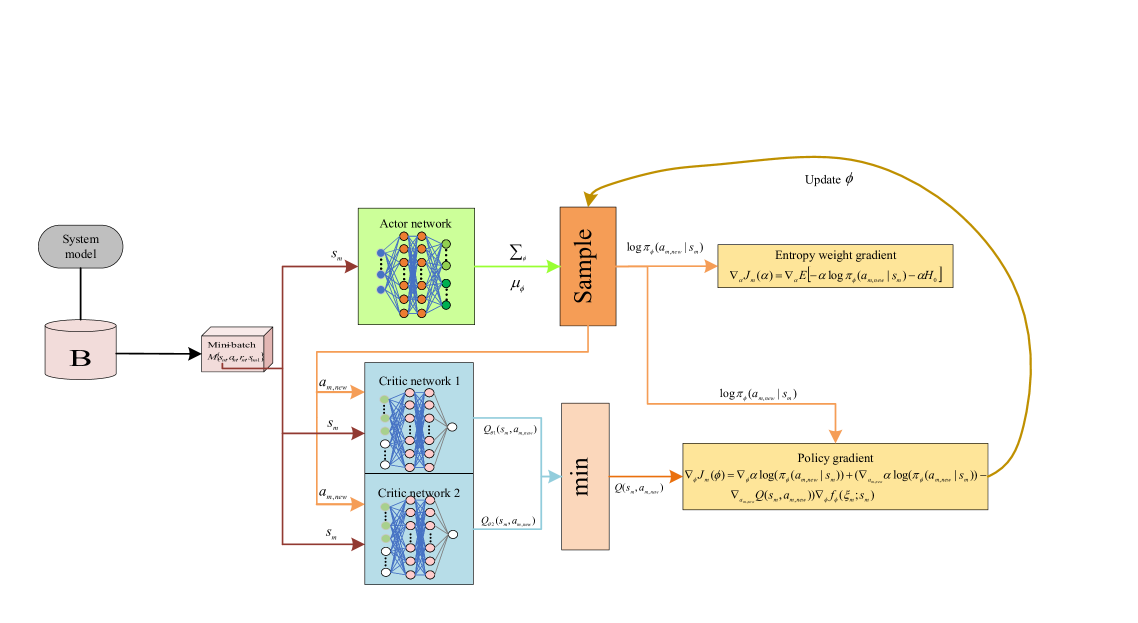}
		\caption{Process to update parameters of actor network}
		\label{fig2}
	\end{figure*}
	\begin{figure*}[t]
		\centering
		\includegraphics[width=5.6in]{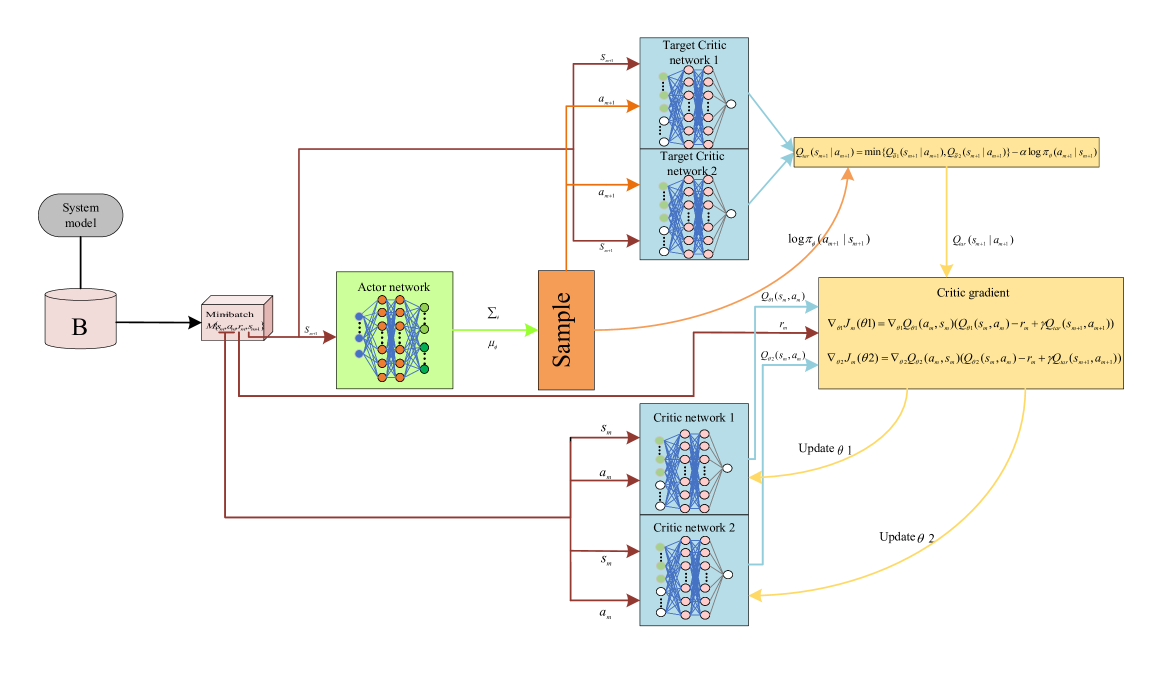}
		\caption{Process to update parameters of the two critic networks}
		\label{fig3}
	\end{figure*}
	
	\subsection{Modeling of the System}
	Now, we first construct this DRL framework with states, actions and rewards and use SAC algorithm to find the optimal policy. The relevant details are described below:
	
	1) \textit{State}: The state of the system at slot $n$ consists of several components: the V2I link local channel,
	$h_m^{c,n}={(h_{r,b}^n[l])^{\rm H}}{\Theta ^n}h_{m,r}^n[l] + h_{m,b}^n[l]$,
	the V2V link local channel,
	$h_k^{d,n}={(h_{r,k}^n[l])^{\rm H}}{\Theta ^n}h_{k,r}^n[l] + h_k^n[l]$, 
	the interference of the previous slot to the receiver in the V2V link,$I_k^{n-1}[l]$, the load left for transmission in the V2V link, $D_k^n$, the AoI of V2I links, $A_m^n$, and the phase-shift matrix of the RIS, $\Theta ^n$. Thus, the state at slot $n$ is given by
	\begin{equation}\label{eq15}
		{s_n} = \left[ {\begin{array}{*{20}{l}}
				{h_1^{c,n},h_2^{c,n}, \cdots ,h_M^{c,n},h_1^{d,n},h_2^{d,n}, \cdots ,h_K^{d,n},}\\
				{I_1^{n - 1},I_2^{n - 1}, \cdots ,I_K^{n - 1},D_1^n,D_2^n, \cdots ,D_K^n,}\\
				{A_1^n,A_2^n, \cdots ,A_M^n,{\Theta ^n},}
		\end{array}} \right].
	\end{equation}
	
	2) \textit{Action}: As mentioned above, the BS acts as an agent that determines the vehicle's channel assignment, power control, and RIS phase-shift matrix, 
	$x_{m,k}^n$, $p_{k,n}^d$ denotes the channel and power selection for the V2V vehicle user pair and ${\Theta ^n}$ denotes the phase-shift selection for the RIS, and thus the action space at slot $n$ is defined as 
	\begin{equation}\label{eq16}
		{a_n} = \left\{ {x_{m,k}^n,p_{k,n}^d,{\Theta ^n}} \right\} .
	\end{equation}
	
	3) \textit{Reward function}: In DRL, rewards are designed to be flexible, and a good reward can improve the performance of the system. In the resource optimization problem described in Section II, our objective has two main components: one is to maintain the connection to the BS and keep the AoI at a minimum level, and the other is to increase the success probability of the V2V efficiently transmitting the load. Therefore, the reward at slot $n$ is set to  
	\begin{equation}\label{eq17}
		{r_n} =  - {\lambda _1}\frac{1}{M}\sum\limits_{m = 1}^M {A_m^n}  - {\lambda _2}\frac{1}{K}\sum\limits_{k = 1}^K {(D_k^n/D)},
	\end{equation}
	where ${\lambda _1}$, ${\lambda _2}$ denotes the weight coefficient used to balance the two parts of reward, the first part denotes the reward obtained for the AoI in the V2I transmission, and the second part denotes the reward obtained for the residual amount of payload transmission in the V2V transmission. 
	
	The next step is to address this problem using the SAC algorithm, which yields the optimal policy. SAC is particularly effective for complex RL tasks due to its policy entropy regularization, experience replay, and dual Q-networks, which enhance stability, sample efficiency, and exploration.
	
	\subsection{Solution}
	\begin{algorithm}[t]
		\caption{AoI-Aware Joint Vehicular Resource Allocation and RIS Phase-Shift Control Scheme Based On SAC Algorithm}
		\label{al1}
		Start environment simulator, generating vehicles and links;
		
		Initialize the $\phi$, $\theta1$, $\theta2$, $\alpha$ randomly;
		
		Initialize target networks by $\overline \theta  1 \leftarrow \theta 1$, $\overline \theta  2 \leftarrow \theta 2$;
		
		Initialize replay experience buffer $\mathcal{B}$;
		
		\For{each episode}
		{Update vehicles locations and respective channel gains;
			
			Reset V2I AoI, V2V payload $D$ and maximum delivery time $N$ to $100$ ms;
			
			\For{each step n}
			{Receive observation state $s_n$;
				
				Generate the RIS phase-shift control policy and vehicular resource allocation policy $a_n$;
				
				Execute action $a_n$, then obtain $r_n$ and next state $s_{n+1}$;
				
				Store tuple $(s_n, a_n, r_n, s_{n+1})$ in replay buffer $\mathcal{B}$;
				
				\If{the size of the replay experience buffer is larger than $I$}
				{Randomly sample a mini-batch of $I$ transitions tuples from $\mathcal{B}$;
					
					Update $\alpha$ by minimizing the loss function according to Eq. (\ref{eq20});
					
					Update the parameter $\phi$ of actor network according to Eq. (\ref{eq25});
					
					Update the parameter $\theta1$ and $\theta2$ of critic network according to Eq. (\ref{eq22}) and Eq. (\ref{eq23});
					
					Update the parameter $\bar \theta1$ and $\bar \theta2$ of target critic network according to Eq. (\ref{eq26}) and Eq. (\ref{eq27}).
					
				}
			}
		}
	\end{algorithm}
	
	In the realm of RIS-assisted vehicular communication networks, the challenge lies in accommodating the high-speed mobility of vehicles and the unavailability of certain environmental information in advance, such as real-time vehicle positions and channel states. Additionally, within the DRL framework, actions and states manifest as high-dimensional continuous variables, presenting a complexity that traditional DRL algorithms needs to address \cite{r35}. To tackle this dynamic decision-making challenge, we introduce an AoI-aware joint scheme for vehicular resource allocation and RIS phase-shift control. This scheme is grounded in the SAC algorithm, which demonstrates proficiency in handling neural networks to approximate high-dimensional states and actions. Firstly, through the maximum entropy objective, SAC not only improves the stability of the strategy, but also makes more effective use of the sample data, thus accelerating the learning process. The combination of its dual Q-learning structure and soft value function further improves the efficiency of policy optimization and reduces the instability of policy update. SAC's relatively robust selection of hyper parameters makes the algorithm perform stably across different tasks and environments, reducing the difficulty of parameter tuning. These features enable SAC to demonstrate strong performance and flexibility when dealing with challenging control tasks. In this subsection, we delve into the training process of SAC.
	
	At each training step $n$, the RL objective is to maximize both the expected reward and the entropy, which can be defined as  
	\begin{equation}\label{eq18}
		J(\pi ({a_n}|{s_n})) = E\left[ {\sum\limits_{n = 0}^N {{\gamma ^{n-1} }{r_{n}}}  + \alpha H({\pi _\theta }(.|{s_n}))} \right],
	\end{equation} where 
	$\gamma $ represents the discount factor, and 
	$H(.)$ is the policy entropy, 
	$\alpha $ is an adjust factor of entropy, controlling the importance of the entropy term, and is known as the temperature parameter, the goal of DRL is to discover the optimal policy 
	\begin{equation}\label{eq19}
		\pi^* = \arg {\max _\pi }J(\pi ({a_n}|{s_n})).
	\end{equation}
	
	During the training process, we save the experience generated from training in buffer $\mathcal{B} = \left[ {{e_1},{e_2}, \cdots ,{e_n}, \cdots } \right]$, When the size of the experience in the buffer is larger than $I$, we will randomly select $I$ tuples from the replay buffer to constitute a minibatch of training data, and let $({s_i},{a_i},{r_i},{s_{i + 1}})$ be the $i$th tuple in the minibatch. Then input $s_i$ into the actor network and get $a_{i,new}$ and 
	$\log {\pi _\theta }({a_{i,new}}|{s_i})$, then we can get the gradient of the loss function with respective to temperature parameter $\alpha$ as 
	\begin{equation}\label{eq20}
		{\nabla _\alpha }{J_i}(\alpha ) = {\nabla _\alpha }E\left[ { - \alpha \log {\pi _\varphi }({a_{i,new}}|{s_i}) - \alpha {H_0}} \right],
	\end{equation} 
	where ${H_0} = \dim ({a_n})$, then input $ s_i$, $a_i$ into two critic networks and output the action-value function 
	${Q_{\theta 1}}({s_i},{a_i})$ and 
	${Q_{\theta 2}}({s_i},{a_i})$, then input $s_{i+1}$ into the actor network and output $a_{i+1}$ and 
	$\log {\pi _\theta }({a_{i + 1}}|{s_{i + 1}})$, and lastly, input $s_{i+1}$ and $a_{i+1}$ into two target critic networks and output 
	${Q_{\bar \theta 1}}({s_{i + 1}}|{a_{i + 1}})$ and 
	${Q_{\bar \theta 2}}({s_{i + 1}}|{a_{i + 1}})$, then we can calculate the target value as follows 
	\begin{equation}\label{eq21}
		\begin{array}{c}
			{Q_{tar}}({s_{i + 1}}|{a_{i + 1}}) = \min \left\{ {{Q_{\bar \theta 1}}({s_{i + 1}}|{a_{i + 1}}),{Q_{\bar \theta 2}}({s_{i + 1}}|{a_{i + 1}})} \right\}\\
			- \alpha \log {\pi _\theta }({a_{i + 1}}|{s_{i + 1}})
		\end{array},
	\end{equation}
	the gradient of the loss function with parameters $\theta1$ and $\theta2$ are as follows
	\begin{equation}\label{eq22}
		\begin{aligned}
			{\nabla _{\theta 1}}{J_i}(\theta 1) &= {\nabla _{\theta 1}}{Q_{\theta 1}}({a_i},{s_i})({Q_{\theta 1}}({s_i},{a_i})\\& - 
			{r_i} + \gamma {Q_{tar}}({s_{i + 1}},{a_{i + 1}})),
		\end{aligned}
	\end{equation}
	\begin{equation}\label{eq23}
		\begin{aligned}
			{\nabla _{\theta 2}}{J_i}(\theta 2) &= {\nabla _{\theta 2}}{Q_{\theta 2}}({a_i},{s_i})({Q_{\theta 2}}({s_i},{a_i})\\& - {r_i} + \gamma {Q_{tar}}({s_{i + 1}},{a_{i + 1}})).
		\end{aligned}
	\end{equation}
	
	Then, the algorithm will then compute the gradient of the 
	$\phi$  loss function by first inputting $s_i$ and $a_{i,new}$ into the two critic networks to get 
	${Q_{\theta 1}}({s_i},{a_{i,new}})$ and ${Q_{\theta 2}}({s_i},{a_{i,new}})$, and then taking the minimum of them to get 
	\begin{equation}\label{eq24}
		Q({s_i},{a_{i,new}}) = \min \left\{ {{Q_{\theta 1}}({s_i},{a_{i,new}}),{Q_{\theta 2}}({s_i},{a_{i,new}})} \right\},
	\end{equation} 
	the gradient of the $\phi$ loss function can then be computed as 
	\begin{equation}\label{eq25}
		\begin{aligned}
			{\nabla _\phi }{J_i}(\phi ) &= {\nabla _\phi }\alpha \log ({\pi _\phi }({a_{i,new}}|{s_i})) \\&+ ({\nabla _{{a_{i,new}}}}\alpha \log ({\pi _\phi }({a_{i,new}}|{s_i})) \\&- {\nabla _{{a_{i,new}}}}Q({s_i},{a_{i,new}})){\nabla _\phi }{f_\phi }({\xi _i};{s_i}),
		\end{aligned}
	\end{equation}
	where $\xi$ is a kind of noise sampled from a multivariate normal distribution, and ${f_\phi }({\xi _i};{s_i})$ is a function of a reparameterized action $a_{i,new}$. 
	
	The process to update parameters of actor network $\phi$ and critic network $\theta1$, $\theta2$ are shown in Figs. \ref{fig2} and \ref{fig3}. In addition, the gradient of the Adam optimizer based on the above loss function is used to update $\alpha$, $\theta1$, $\theta2$, $\phi$. Finally, the parameters 
	$\bar \theta 1$, $\bar \theta 2$ of the two target critic networks are updated to 
	\begin{equation}\label{eq26}
		\bar \theta 1 = \tau_1\theta 1 + (1 - \tau_1)\bar \theta 1,
	\end{equation}
	\begin{equation}\label{eq27}
		\bar \theta 2 = \tau_2\theta 2 + (1 - \tau_2)\bar \theta 2,
	\end{equation} 
	where $\tau_1$ and $\tau_2$ are constants that satisfy 
	$\tau_1 \ll 1$ and $\tau_2 \ll 1$.
	After all the parameters are updated, the algorithm will continue to the next segment and when the training is over, the algorithm will get the optimal policy. The training procedure is summarized in Algorithm \ref{al1}.
		
	\subsection{Computational Complexity Analysis}
	In this section, we will analyze the computational complexity of our approach. We first analyze the computational complexity during the training stage. Since training requires significant computational resources to compute gradients and update parameters, the overall complexity consists of the complexity associated with gradient computation and parameter updates. Let $G_E$, $G_A$, and $G_C$ represent the computational complexity for computing gradients related to the policy entropy tradeoff weight $\alpha$, the actor network $\phi$, and the two critic networks $\theta1$ and $\theta2$, respectively. Similarly, let $U_E$, $U_A$, and $U_C$ denote the computational complexity for updating the parameters of $\alpha$, $\phi$, $\theta1$, and $\theta2$, respectively. Note that the two target critic networks have the same structure as the critic networks and only require updates to their parameters, the target networks have the same complexity of parameter updating as critic networks.  We assume $E$ is the algorithm training episode, and each episode contains $S$ time slots for training, in addition, parameter updates and gradient computations are executed only when the tuples stored in the reply buffer is greater than $I$. Therefore, the complexity of our approach in the training stage is $\mathcal{O}\left( {(E \cdot S - I)({G_E} + {G_A} + 2{G_C} + {U_E} + {U_A} + 4{U_C})} \right)$.
	
	\section{Simulation Results}
	\begin{table}[t]
		\setlength{\tabcolsep}{-0.5pt}
		\begin{center}
			\caption{Simulation Parameters.}
			\label{tab1}
			\begin{tabular}{|c|c|}
				\hline
				\textbf{Parameter} & \textbf{Value}\\
				\hline
				Number of V2I links $M$ & 4\\
				\hline
				Number of V2V links $K$ & 4 \\
				\hline
				Phase-shift variable quality $Q$ & 8 \\
				\hline
				Number of RIS elements $F$ & 12 \\
				\hline
				Carrier frequency & 2 GHz\\
				\hline
				Bandwidth & 1 MHz \\
				\hline
				BS antenna height $z_{BS}$ & 25 m \\
				\hline
				RIS height $z_{RIS}$ & 25 m \\
				\hline
				Vehicle antenna height & 1.5 m \\
				\hline
				BS and vehicles antenna gains & 8, 3 dBi \\
				\hline
				BS and vehicles receiver noise gains & 5, 11 dBi \\
				\hline
				Noise power ${\sigma ^2}$ & -114 dBm \\
				\hline
				V2I transmit power $P^c$ & 23 dBm \\
				\hline
				V2V transmit power $P^d$ & $[1,2,\cdots,23]$ dBm\\
				\hline 
				V2I link minimum transmission rate $R^{th}$ & 3 bps/Hz\\
				\hline
				Vehicle speed range & $[10,15]$ m/s \\
				\hline
				\makecell[c]{Time constraint of V2V payload\\ transmission $N$ } & 100 ms\\
				\hline
				V2V initial payload size $D$ & 8 $\times$ 1060 bytes\\
				\hline
				weight coefficient $\lambda_1$, $\lambda_2$ & 10, 1\\
				\hline
				V2I link path loss model & $128.1 + 37.6log_{10}(d)$\\
				\hline
				V2V link path loss model & \makecell[c]{LOS in WINNER +\\ B1 Manhattan}\\
				\hline
				\makecell[c]{Shadowing standard deviation for\\ V2I and V2V links} & 8 dB, 3dB\\
				\hline
				\makecell[c]{Decorrelation distance for V2I and\\ V2V links} & 50 m, 10 m\\
				\hline
				Fast fading & Rayleigh fading\\
				\hline
				\makecell[c]{Path loss and shadowing update for\\ V2I and V2V links} & Every 100 ms\\
				\hline
				\makecell[c]{Fast fading update for V2I and V2V links} & Every 1 ms\\
				\hline
			\end{tabular}
		\end{center}
	\end{table}
	
	\begin{table}[t]
		\begin{center}
			\caption{Neural Networks Parameters}
			\label{tab2}
			\begin{tabular}{|c|c|}
				\hline
				\textbf{Parameter} & \textbf{Value}\\
				\hline
				Optimizer & Adam\\
				\hline
				Discount factor $\gamma$ & 0.99 \\
				\hline
				Critic/Actor networks learning rate & $3 \cdot {10^{ - 4}}$\\
				\hline
				Nonlinear function & Relu \\
				\hline
				Experience replay buffer size $\mathcal{B}$ & $10^6$\\
				\hline
				Mini-batch size $I$ & $10^6$\\
				\hline
				Number of episodes & 1000\\
				\hline
				Number of iterations per episode & 100\\
				\hline
				Size of hidden layers & 512\\
				\hline
				Target network soft update parameters, $\tau_1$, $\tau_2$ & 0.01\\
				\hline
			\end{tabular}
		\end{center}
	\end{table}
	\begin{figure}[t]
		\centering
		\includegraphics[width=3.4in]{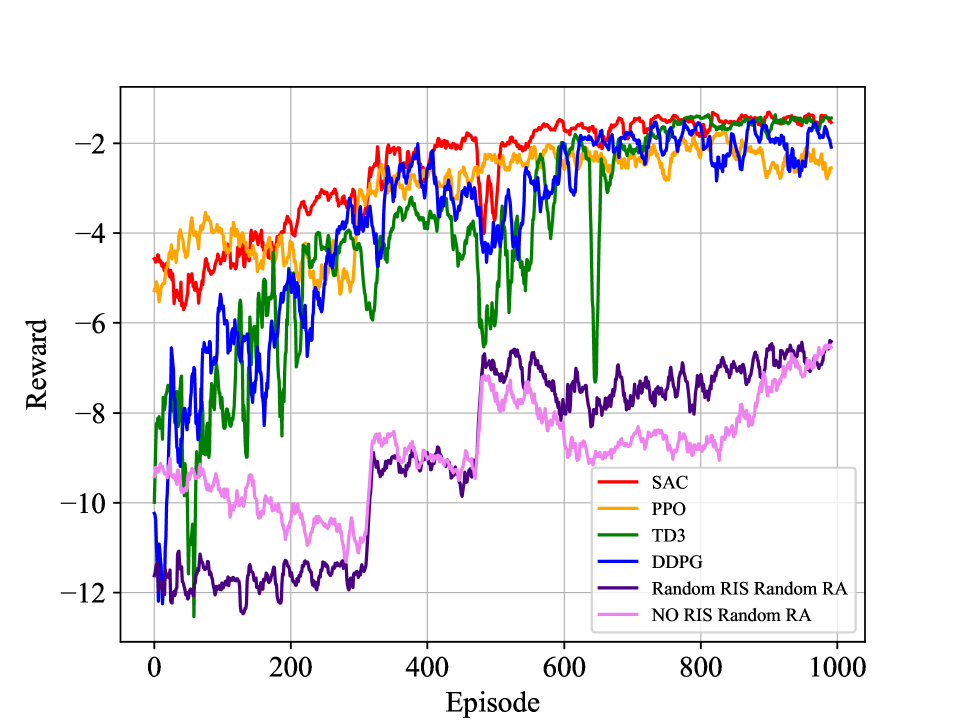}
		\caption{Reward comparison}
		\label{fig4}
	\end{figure}
	
	In this section, we use simulation to check the performance of our the proposed centralized vehicular network decision scheme based on SAC algorithm for AoI-aware joint spectrum, power, and phase-shift. The simulation tool is python $3.9$. We construct the vehicular environment according to \cite{r36}, \cite{r46} and \cite{r45}, and list the relevant parameters in Tables \ref{tab1} and \ref{tab2}. The three-dimensional coordinates of the BS and the RIS are (180m, 270m, 25m) and (290m, 380m, 25m) respectively. $M$ vehicles form $M$ V2I links, and $K$ V2V links are formed by the vehicles and their neighboring vehicles, where all the links are formed by the direct and RIS-assisted reflective links. At intersections, each vehicle has a fixed probability of choosing a turn, here we set it to 0.4, and in addition, the initial vehicle speed is randomly chosen from the interval $[10m/s, 15m/s]$. We set the V2V initial transmission payload size to 8*1060 bits and the V2I link initial AoI to 100 ms units during training.
	
	The structure of the SAC algorithm consists of an actor network, two critic networks, and two target critic networks, all of which consist of a five-layer fully connected neural network including one input layer, one output layer, and three hidden layers, where the number of neurons in each of the three hidden layers is 512. During the training process, we set up 1000 training episodes, and the discount factor $\gamma $ is set to 0.99. At the end of the training, we conduct tests based on the training model, each round contains 50 test slots, and each result is obtained by averaging the results of sequential 10 tests.
	
	We compare the proposed SAC algorithm with the following benchmarks:
	\begin{itemize}
		\item [1)] \textbf{Traditional centralized algorithms}: In this group we will use three different algorithms, PPO, TD3, and DDPG to control the spectrum and power of the vehicle as well as the phase-shift of the RIS.
		
		\item [2)] \textbf{Random RIS Random RA}: This scheme sets both the phase-shift matrix of the RIS and the resource allocation of the vehicle to be random.
		
		\item [3)] \textbf{NO RIS Random RA}: In this group, we will not use RIS for auxiliary communication with the vehicle, and in addition, we will randomize the allocation of the vehicle spectrum and power.
	\end{itemize}

	In Fig. \ref{fig4}, the training performance of the proposed SAC algorithm is showcased alongside several baseline schemes, focusing on rewards as the performance metric. The visual representation reveals that the SAC algorithm exhibits superior performance with higher reward values, demonstrating faster and more stable convergence compared to the alternative baseline schemes. Notably, when confronted with high-dimensional state spaces, both DDPG and TD3 exhibit initial training challenges. The distinctive advantage of the SAC algorithm becomes apparent, showcasing its efficacy in addressing the complexities posed by high-dimensional state spaces and ensuring efficient and stable convergence throughout the training process.
	
	\begin{figure}[t]
		\centering
		\includegraphics[width=3.4in]{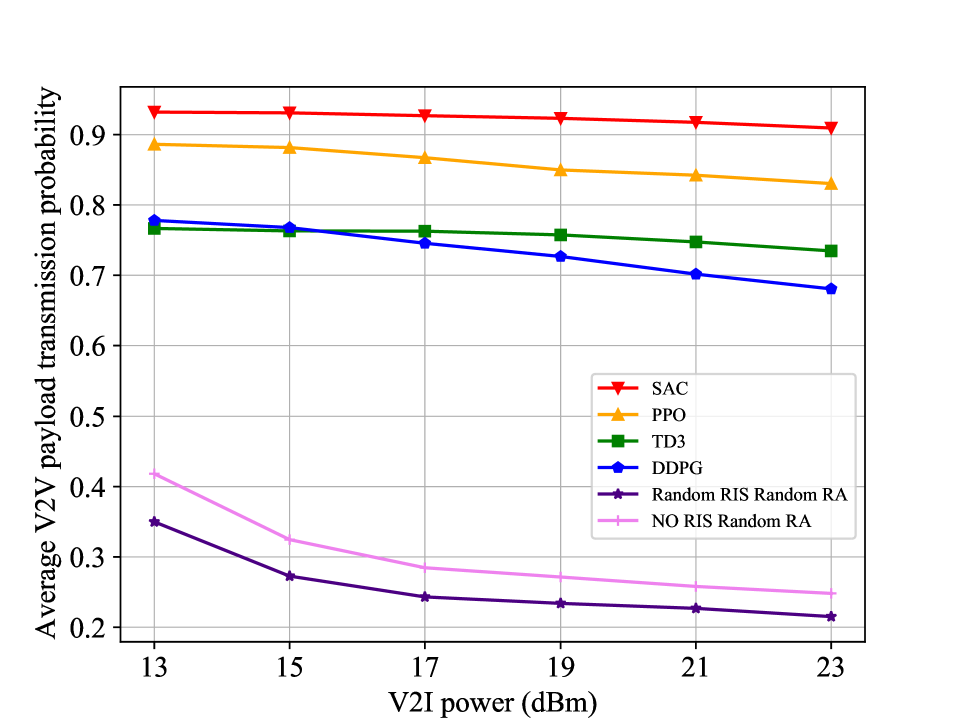}
		\caption{V2V payload transmission success probability with varying V2I power}
		\label{fig5}
	\end{figure}
	
	\begin{figure*}[htbp]
		\centering
		\subfloat[]
		{\includegraphics[width=3.4in]{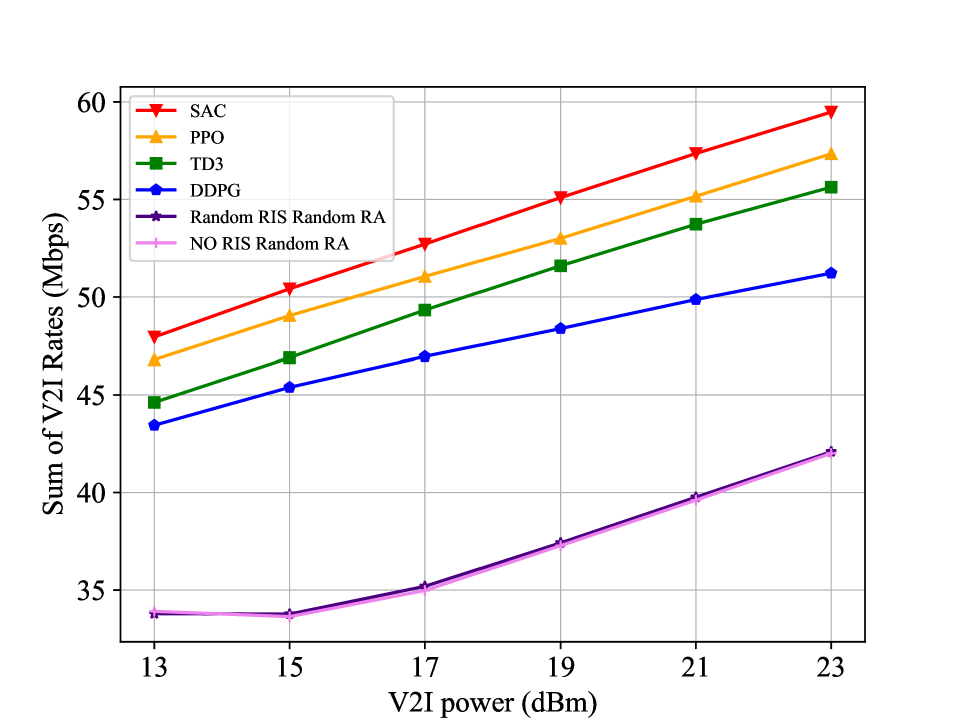}}
		\hfil
		\subfloat[]
		{\includegraphics[width=3.4in]{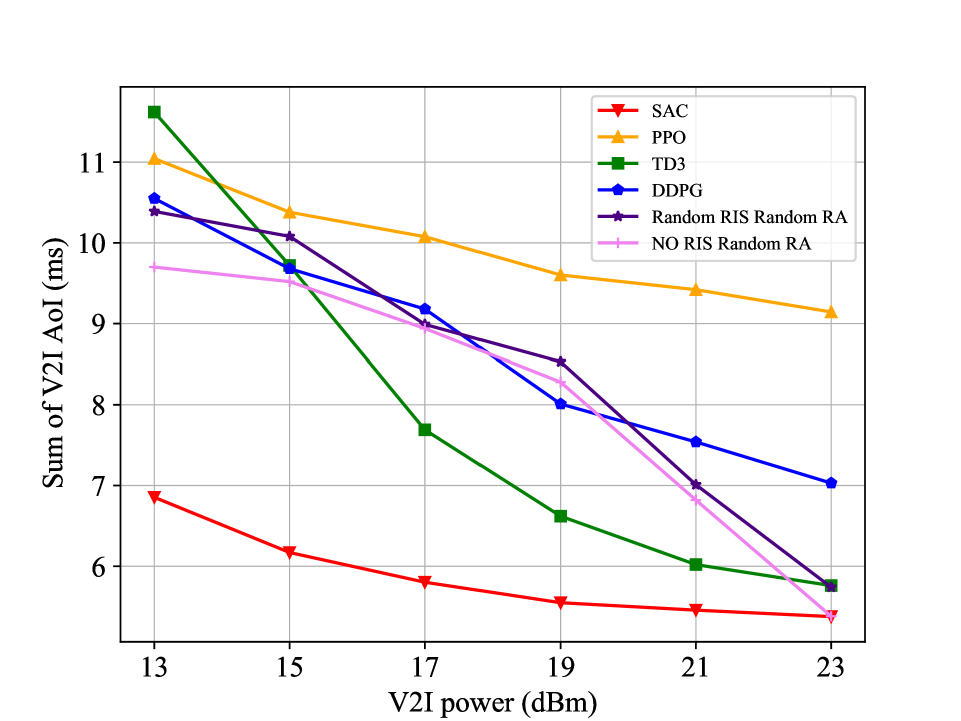}}
		\caption{Performance versus V2I power. (a) Sum of V2I rates, (b) Sum of V2I AoI.}
		\label{fig6}
	\end{figure*}
	
	In order to have verified the reliability of V2V transmission, we expressed it by the payload transmission success rate of V2V. Fig. \ref{fig5} sheds light on the relationship between V2I transmission power and V2V payload transmission success probability. The observed decrease in success probability is attributed to the amplified interference caused by higher V2I transmission power. This interference adversely affects the information rate of the V2V link, consequently impeding the V2V links' ability to complete information transmission within the allotted time, resulting in a decline in the success rate. When the V2I power is increased from 13 dBm to 23 dBm, the payload transmission success rate of V2V under the SAC algorithm decreases by 2.43$\%$. In addition, when the V2I power is 23 dBm, the payload transmission success rate of V2V under the SAC algorithm is improved by 33.2$\%$ compared to DDPG.
	
	In Fig. \ref{fig6}, we conduct a comparative analysis of various methods, assessing their performance in terms of V2I AoI and V2I transmission rate as the V2I power undergoes testing.
	Figs. \ref{fig6}(a) and \ref{fig6}(b) illustrate the impact of increasing V2I transmission power on the total rate of V2I links and the corresponding AoI. As the V2I transmission power rises, the SINR at the BS increases, leading to an elevation in the transmission rate. This increase in power results in more time slots surpassing the threshold, contributing to a reduction in AoI. When the V2I power is increased from 13 dBm to 23 dBm, the V2I information transmission rate under the SAC algorithm is improved by 24.04$\%$ and the AoI of V2I is reduced by 21.56$\%$.
	
	\begin{figure*}[htbp]
		\centering
		\subfloat[] 
		{\includegraphics[width=3.4in]{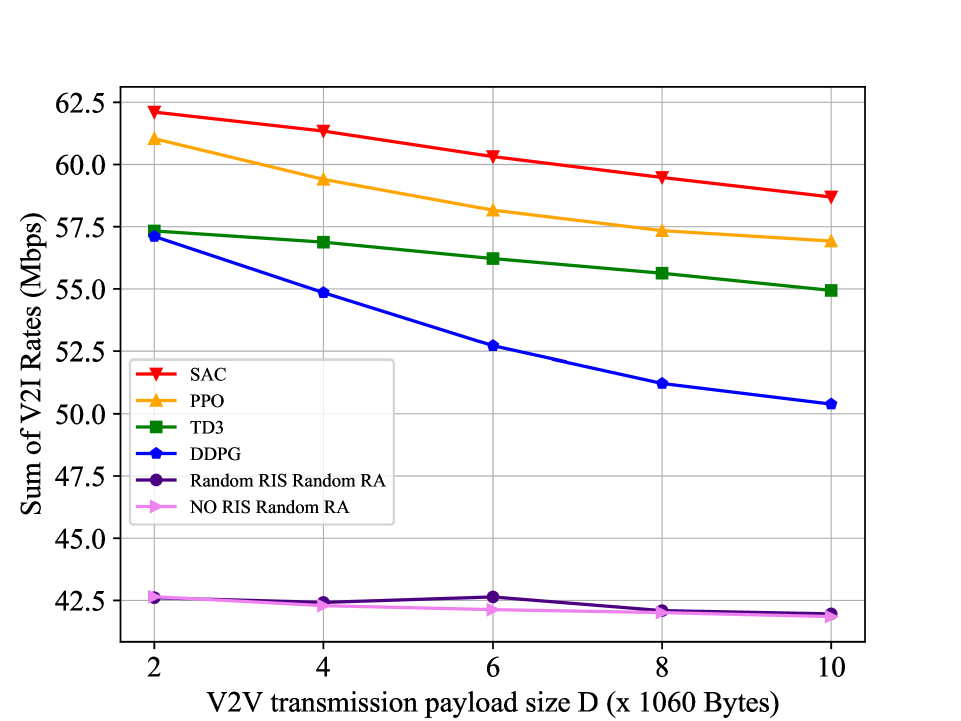}}
		\hfil
		\subfloat[]
		{\includegraphics[width=3.4in]{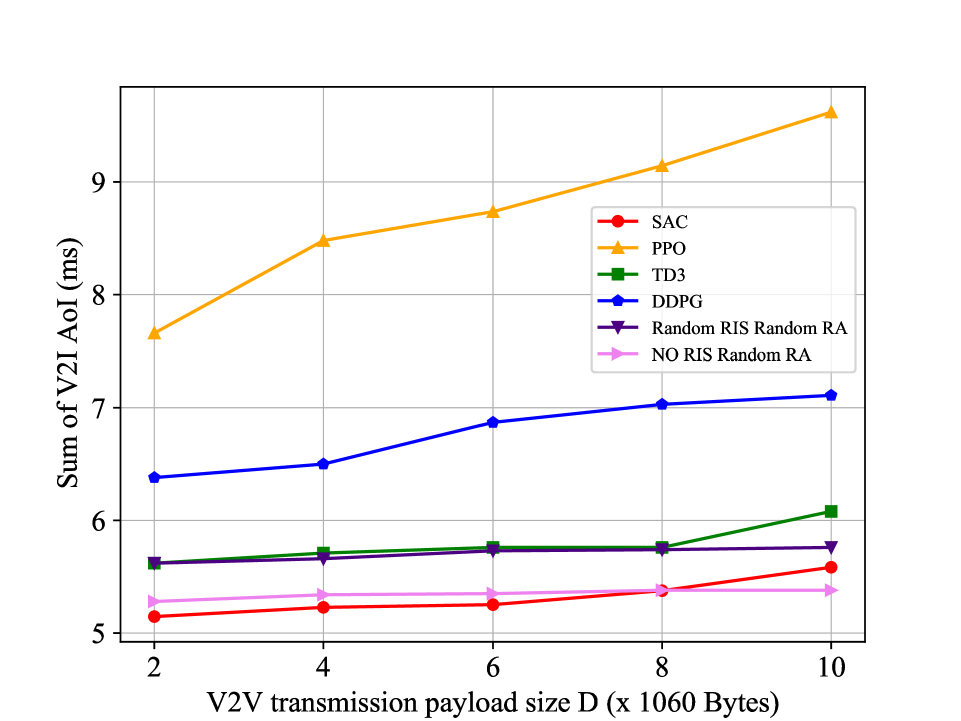}}
		\caption{Performance versus V2V payload $D$. (a) Sum of V2I rates, (b) Sum of V2I AoI.}
		\label{fig7}
	\end{figure*}
	
	In Fig. \ref{fig7}, the impact of the V2V link payload size on both the V2I link information transmission rate and AoI is illustrated. Fig. \ref{fig7}(a) demonstrates that the information transmission rate of all schemes experiences a decline with an increase in the V2V payload size. Correspondingly, Fig. \ref{fig7}(b) reveals a gradual rise in AoI as the V2V payload size increases. The decrease in information transmission rate is attributed to the elevated V2V payload size, resulting in a prolonged V2V transmission time. This extension may necessitate higher transmission power for the V2V link, subsequently intensifying interference with the V2I link. Moreover, the prolonged V2V transmission time perpetuates the continuous sharing of V2I link spectrum resources by the V2V link. This persistent interference from the V2V link to the V2I link diminishes the information transmission rate, contributing to the gradual increase in AoI. When the V2V transmission payload size is increased from 2120 bytes to 10600 bytes, the V2I information transmission rate under the SAC algorithm decreases by 5.5 $\%$ and the AoI of V2I link increases by 8.51$\%$. The observed dynamics underscore the critical role of V2V payload size considerations in influencing the performance trade-offs within the vehicular communication network.

	\begin{figure*}[htbp]
		\centering
		\subfloat[] 
		{\includegraphics[width=3.4in]{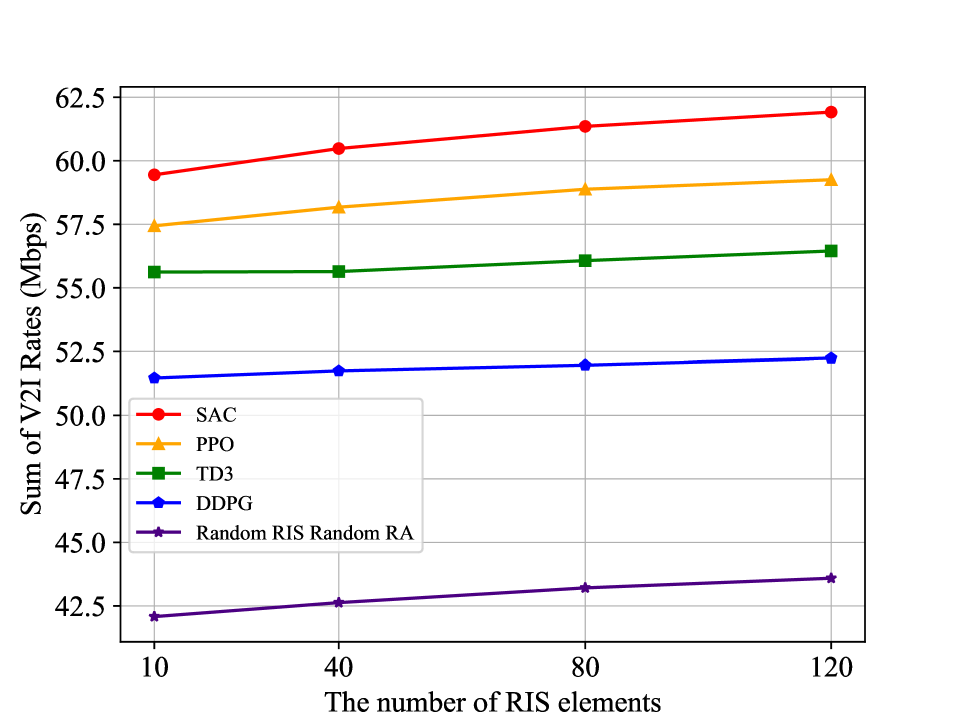}}
		\hfil
		\subfloat[]
		{\includegraphics[width=3.4in]{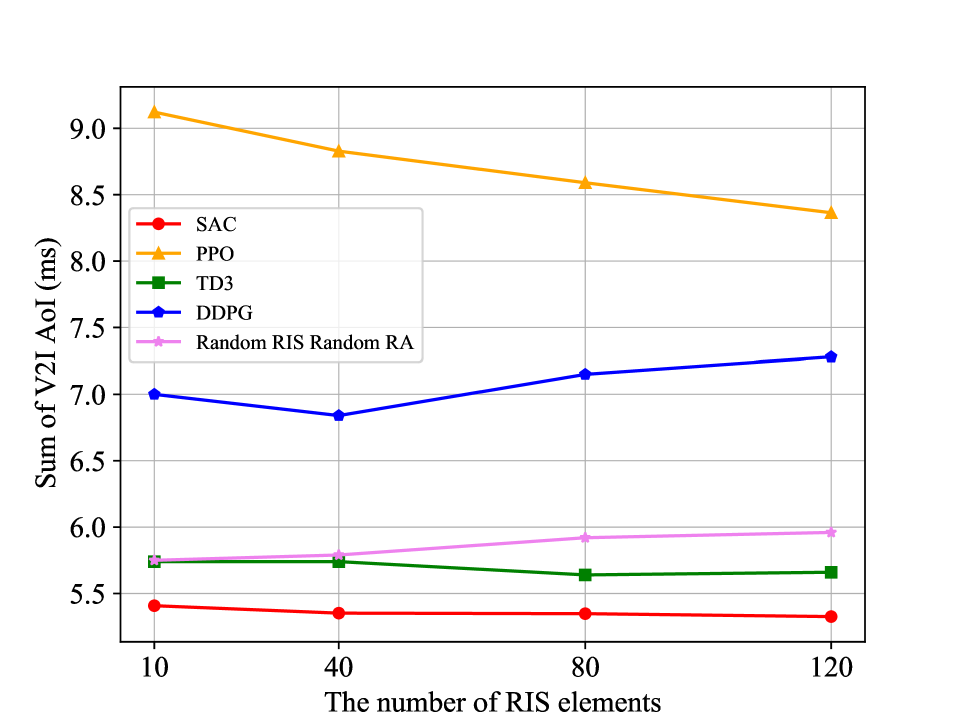}}
		\caption{Performance versus the number of RIS. (a) Sum of V2I rates, (b) Sum of V2I AoI}
		\label{fig8}
	\end{figure*}
	
	\begin{figure}[htbp]
		\centering
		\includegraphics[width=3.4in]{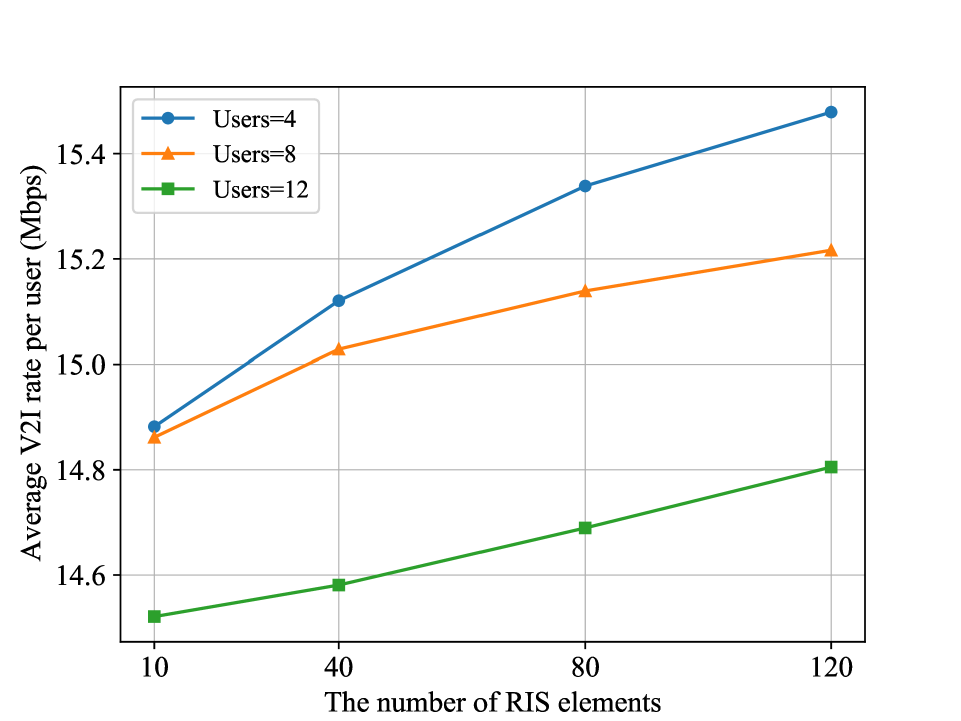}
		\caption{Average rate per user over the number of RIS elements with different number users}
		\label{fig9}
	\end{figure}
	Furthermore, Fig. \ref{fig8} illustrates the impact of varying the number of RIS on V2I information transmission rate and AoI. When the number of RIS elements is increased from 10 to 120, the V2I information transmission rate under the SAC algorithm is improved by 4.15 $\%$ and the AoI of V2I is reduced by 1.52$\%$. This observation provides validation for the beneficial influence that RIS imparts to the overall system performance. The augmented number of RIS enhances the likelihood of signals being reflected through the RIS to the receivers, contributing to improved V2I performance. However, it's essential to note that as the number of RIS increases, so does the interference among users. Additionally, the heightened complexity introduced by the increased number of RIS needs to be carefully considered. Hence, the deployment of RIS becomes a crucial factor that necessitates thoughtful evaluation during model training. The trade-off between the advantages and complexities introduced by RIS deployment underscores the importance of optimizing the number of RIS for enhanced system performance.
	
	To assess the impact of an increasing number of users on system performance, we examine the variation in the average information transmission rate for each user as the number of RIS increases. Specifically, we observe this change with varying user counts, including scenarios with 4, 8, and 12 users. Fig. \ref{fig9} illustrates that the average information transmission rate for each user exhibits an upward trend with the increasing number of RIS. Notably, in situations with a larger number of users, the average information transmission rate for each user tends to decrease. This is attributed to the heightened interference among users, a consequence of the increased user count. The rise in interference reduces the SINR, ultimately resulted in a decline in the information transmission rate for individual users. This observation emphasizes the importance of considering the trade-off between the benefits gained from RIS deployment and the challenges introduced by an escalating number of users in optimizing overall system performance.
	
	In summary, the comprehensive comparison tests conducted reveal that our proposed SAC algorithm holds a distinct advantage across various performance metrics, including information transmission rate and AoI. The SAC algorithm emerges as a leading-edge solution, showcasing superior performance in optimizing the efficiency and effectiveness of the vehicular communication network. These findings underscore the efficacy of the SAC algorithm in addressing the challenges posed by high-dimensional state spaces and dynamic decision-making scenarios, making it a promising approach for enhancing the overall performance of the proposed vehicular network decision scheme.
	
	\section{Conclusion}
	In this study, we focused on enhancing V2X communication performance by considering a RIS-assisted vehicular communication environment. Our objective is to minimize the AoI of the V2I link and increase the payload delivery success rate of the V2V link. To achieve this optimization goal, we formulated the problem as a single-agent MDP using DRL and proposed an AoI-aware joint scheme for vehicular resource allocation and RIS phase-shift control based on the SAC algorithm. Extensive simulations with different parameters demonstrated the effectiveness of the proposed method and showed that it outperforms other algorithms such as DDPG, PPO, and TD3 in terms of AoI, information transmission rate, and payload transmission success rate. The conclusions are summarized as follows:
	\begin{itemize}
		\item [\textbullet] Our approach supports real-time decision making, and the SAC algorithm performs well when dealing with high-dimensional state spaces, coping effectively with complexity and ensuring efficient and stable convergence of the training process.
		
		\item [\textbullet] The deployment of RIS and the optimization of the RIS phase-shift matrix positively affect the overall system performance, with both the V2I information transfer rate and AoI improving as the number of RIS increases, highlighting its potential benefits.
		
		\item [\textbullet] Reasonable consideration and adjustment of the V2V load size and V2I transmission power help to balance the performance in the communication network and improve the information transmission efficiency.
	\end{itemize}

\end{document}